\documentclass[letterpaper, 10 pt, conference]{ieeeconf}  % Comment this line out if you need a4paper

\IEEEoverridecommandlockouts                              % This command is only needed if 
\usepackage{graphicx} % for pdf, bitmapped graphics files
\usepackage{amsmath} % assumes amsmath package installed
\usepackage{amssymb}  % assumes amsmath package installed
\usepackage{amsbsy}

\usepackage{todonotes}
\usepackage{verbatim}

\usepackage{algorithm}% http://ctan.org/pkg/algorithms
\usepackage{algpseudocode}% http://ctan.org/pkg/algorithmicx

\usepackage[utf8]{inputenc}

\usepackage{cite}
\usepackage{subcaption}
\usepackage{booktabs}
\usepackage{tikz}
\usepackage{pgfplots}

\pgfplotsset{grid style={dotted, gray}}
\pgfplotsset{minor grid style={dotted,gray}}
\pgfplotsset{every tick label/.append style={font=\tiny}}
\pgfplotsset{every axis/.append style={font=\small}}
\pgfplotsset{ylabel near ticks}
\pgfplotsset{xlabel near ticks}
\newlength\figureheight 
\newlength\figurewidth

\title{\LARGE \bf
Compliant Manipulation of Free-Floating Objects
}

\author{Shikha Sharma, Markku Suomalainen, and Ville Kyrki% <-this % stops a space
\thanks{This work was supported by Academy of Finland, decision 286580.}% <-this % stops a space
\thanks{S.\ Sharma, M.\ Suomalainen and V.\ Kyrki are with School of Electrical Engineering, Aalto University, Finland, \tt\small first.last@aalto.fi}}

\begin{document}

\maketitle
\thispagestyle{empty}
\pagestyle{empty}

%%%%%%%%%%%%%%%%%%%%%%%%%%%%%%%%%%%%%%%%%%%%%%%%%%%%%%%%%%%%%%%%%%%%%%%%%%%%%%%%
\begin{abstract}
Compliant motions allow alignment of workpieces using naturally occurring interaction forces. However, free-floating objects do not have a fixed base to absorb the reaction forces caused by the interactions. Consequently, if the interaction forces are too high, objects can gain momentum and move away after contact. This paper proposes an approach based on direct force control for compliant manipulation of free-floating objects. The objective of the controller is to minimize the interaction forces while maintaining the contact. The proposed approach achieves this by maintaining small constant force along the motion direction and an apparent reduction of manipulator inertia along remaining Degrees of Freedom (DOF). Simulation results emphasize the importance of relative inertia of the robotic manipulator with respect to the free-floating object. The experiments were performed with KUKA LWR4+ manipulator arm and a two-dimensional micro-gravity emulator (object floating on an air bed), which was developed in-house. It was verified that the proposed control law is capable of controlling the interaction forces and aligning the tools without pushing the object away. We conclude that direct force control works better with a free-floating object than implicit force control algorithms, such as impedance control. 

\end{abstract}

%%%%%%%%%%%%%%%%%%%%%%%%%%%%%%%%%%%%%%%%%%%%%%%%%%%%%%%%%%%%%%%%%%%%%%%%%%%%%%%%
\section{INTRODUCTION}
\label{intro}

Manipulation has emerged as one of the major fields in robotics research in the past few decades \cite{HandbookRobotics}. Typical examples of manipulation tasks are assembly of objects and alignment of workpieces. These tasks involve interaction with the object, thus requiring compliance. Compliance can also mitigate position uncertainties of objects being aligned. In addition to compliance, control of interaction forces between objects is required for successful completion of these tasks. Therefore, the controller plays an important role in manipulation operations and its goal is to successfully control the motion as well as contact forces.

\begin{figure}[!ht]
  \begin{center}
    \includegraphics[width=0.7\linewidth]{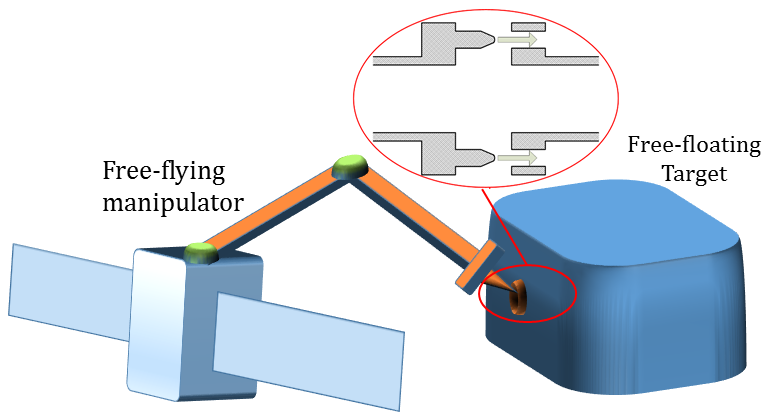}
    \caption{Sketch showing compliant manipulation (alignment of workpieces) in free-floating environment}
    \label{fig:schematic_free}
  \end{center}
\end{figure}

In space, underwater, or aerial applications, the manipulator and the target are free-floating such that a fixed base does not absorb reaction forces (Fig. \ref{fig:schematic_free}). If the interaction forces are too high, the object can gain momentum and move away after contact. Additionally, the motion of the object after contact depends on the relative inertia of the end-effector with respect to the object. Hence the objective of a controller is to minimize the interaction forces along with minimizing relative inertia while maintaining the contact. Maintaining contact is important, contact breaking results in an impact force every time the bodies come in contact during the manipulation task.  

Impedance control is one of the most popular methods for compliant manipulation, in which the controller imposes mechanical impedance of an equivalent mass-spring-damper system with adjustable parameters on the end-effector. However, it is an implicit way of force control, i.e. it indirectly regulates the contact forces by generating an appropriate motion that ends up in a desired dynamic relationship between the robot and the environment \cite{HoganImpedance1985}. Moreover, if the object is free-floating, there is no control over its position. Hence it is difficult to maintain a desired value of force with impedance control especially in context of free-floating object. 

In this paper, we propose an approach which maintains a constant force along the motion direction and apparent reduction of manipulator inertia similar to virtual tool along remaining DOF. The proposed approach is based on direct force control for compliant manipulation while minimizing the contact force. Simulation results are presented, emphasizing the importance of relative inertia between the manipulator and free-floating object when motion is constrained along one dimension. The approach is studied experimentally with an inexpensive two-dimensional micro-gravity emulator setup developed in-house and a 7-DOF manipulator arm. The object floating on an air bed is used to verify the proposed control algorithm for maintaining minimum interaction force and to achieve alignment of tools. Experiments demonstrate that the direct force control works better with free-floating objects than indirect force control.

Section II reviews related work for compliant manipulation of free-floating objects. Section III discusses simulation and corresponding results. In Section IV we explain the proposed method used for compliant manipulation of free-floating objects. Next, experiments with a KUKA LWR4+ robot arm and their results are presented in Section V. Finally, the results are discussed and future work is outlined in Section VI.

\section{RELATED WORK}
\label{RELATED}

Compliant motions have many advantages in manipulation tasks, such as assembly or alignment of tools. Compliant manipulation can be achieved by controlling interaction forces passively or actively. In passive interaction control, the inherent compliance of the robot is exploited, e.g., structural compliance of links, joints, and end-effector. On the other hand, active interaction control ensures the compliance by a purposely designed control system. In practical robot systems some combination of active and passive interaction control is often employed \cite{Chen2013,Schutter1988}.

A popular approach for compliant manipulation is impedance control. In impedance control the end-effector deviation from prescribed trajectory due to environment gives rise to contact forces \cite{HoganImpedance1985}. Variations of impedance control and other simplified control strategies have also been used, such as admittance control, stiffness control, damping control \cite{Whitney1977}, and compliance control \cite{SalisburyStiffness1980}. Additionally, direct force control approaches such as hybrid force-motion control have also been used for controlling end-effector \cite{RaibertHybrid1981,Khatib1987,Villani1999}. These approaches regulate the contact force to a desired value with an explicit force feedback loop. Most of the work reported for compliant manipulation using impedance or force control of a manipulator arm is related to ground based robots \cite{suomalainen2017}.

Little work has been done for compliant manipulation of free-floating objects.  Some research is available for space manipulators, i.e. manipulator in free-floating environment. Approaches based on compliance control of space manipulators for on-orbit interaction are, for instance, the impedance control of a free-flying space robot proposed by Yoshida, et al. \cite{Yoshida2004, Nakamishi2010, Yoshida2011}, the joint compliance control by Nishida, et al. \cite{Nishida2003}, and impedance control of dexterous space manipulators by Colbaugh, et al. \cite{Colbaugh1992}. As these space manipulators are free-floating, these control approaches are similar to our approach. However, none of the approaches use compliant motions for alignment of workpieces. 

\section{SIMULATIONS}
\label{SIMULATIONS}
\begin{figure}[ht]
  \begin{center}
    \includegraphics[width=0.7\linewidth]{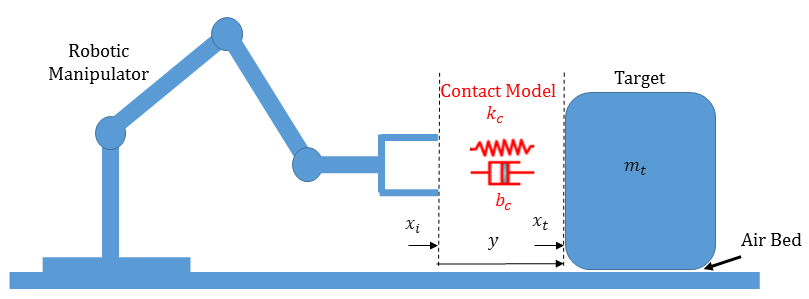}
    \caption{Schematic model of a robotic manipulator arm and free-floating object}
    \label{fig:schematic_sim}
  \end{center}
\end{figure}

Control of interaction forces is central for compliant manipulation. The goal of simulations is to understand the influence of inertia, stiffness and damping on interaction forces. Fig. \ref{fig:schematic_sim} shows the schematic model of a robotic manipulator arm and a free-floating object (on the air bed) with 3-DOF. However, the simulations were performed for 1-DOF in motion.   

The robotic arm is fixed on the table and the end-effector is controlled to have desired compliance characteristics. The system of differential equations for manipulator, object, modeled contact force, and relative acceleration between manipulator and object in one dimension can be written as

\begin{equation} \label{control_sim_man}
\begin{gathered} 
m_{ri}\ddot{x}_i + b_{ri}\Delta \dot{y}+k_{ri}\Delta y =-F_{c} \\
m_{t}\ddot{x}_t =F_{c}-F_{f} \\
F_{c}=-k_cy-b_c\dot{y} \\
\ddot{y} = \ddot{x}_t - \ddot{x}_i \\
\end{gathered}
\end{equation} 

\noindent
where $m_{ri}$ is mass, $b_{ri}$ is damping, and $k_{ri}$ is stiffness. In the simulations, stiffness and damping of the manipulator correspond to passive compliance of the arm. $x_i$ is the position of the manipulator end-effector and $x_t$ the position of the free-floating object. Mass of the free-floating object is $m_{t}$, and $F_{f}$ is force of friction (value used as zero in case of air bed). $F_c$ is contact force exerted on manipulator, which is modeled as a continuous function of penetration and rate of penetration of one rigid body into another. The penetration and rate of penetration are the relative position, $\Delta y = x_t-x_i$, and relative velocity, $\Delta \dot{y}$, respectively, defined between the manipulator hand and a contact point on the target surface. The parameters $k_c$ and $b_c$ are the stiffness and damping of actual contact surfaces, respectively.

\begin{figure*}[ht]
    \centering

    \makebox[\linewidth][c]{
    \begin{subfigure}{0.3\textwidth} \centering \includegraphics[width=1\textwidth]{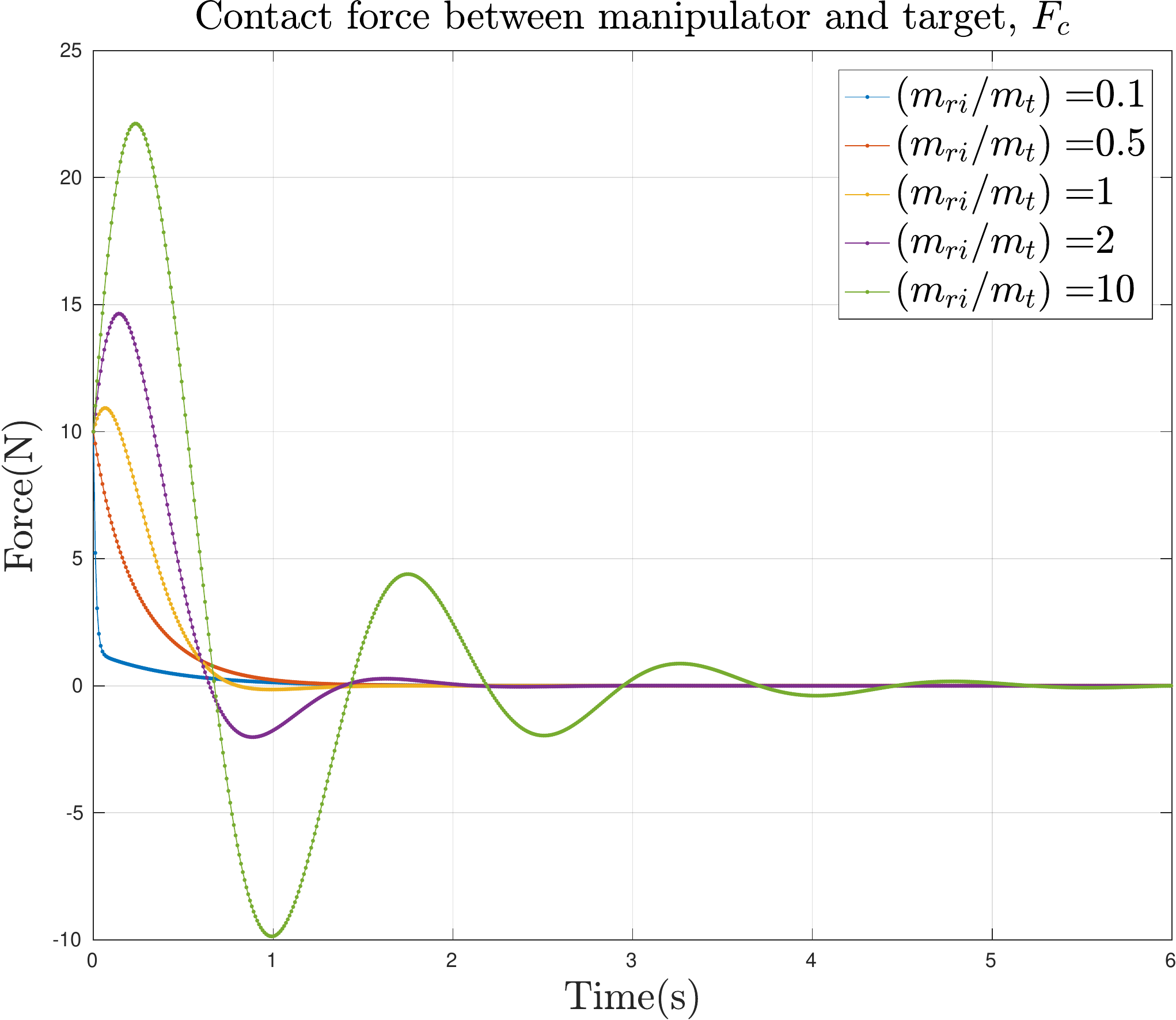} \caption{} \label{fig:sim_res_a} \end{subfigure} \quad \begin{subfigure}{0.3\textwidth} \centering \includegraphics[width=1\textwidth]{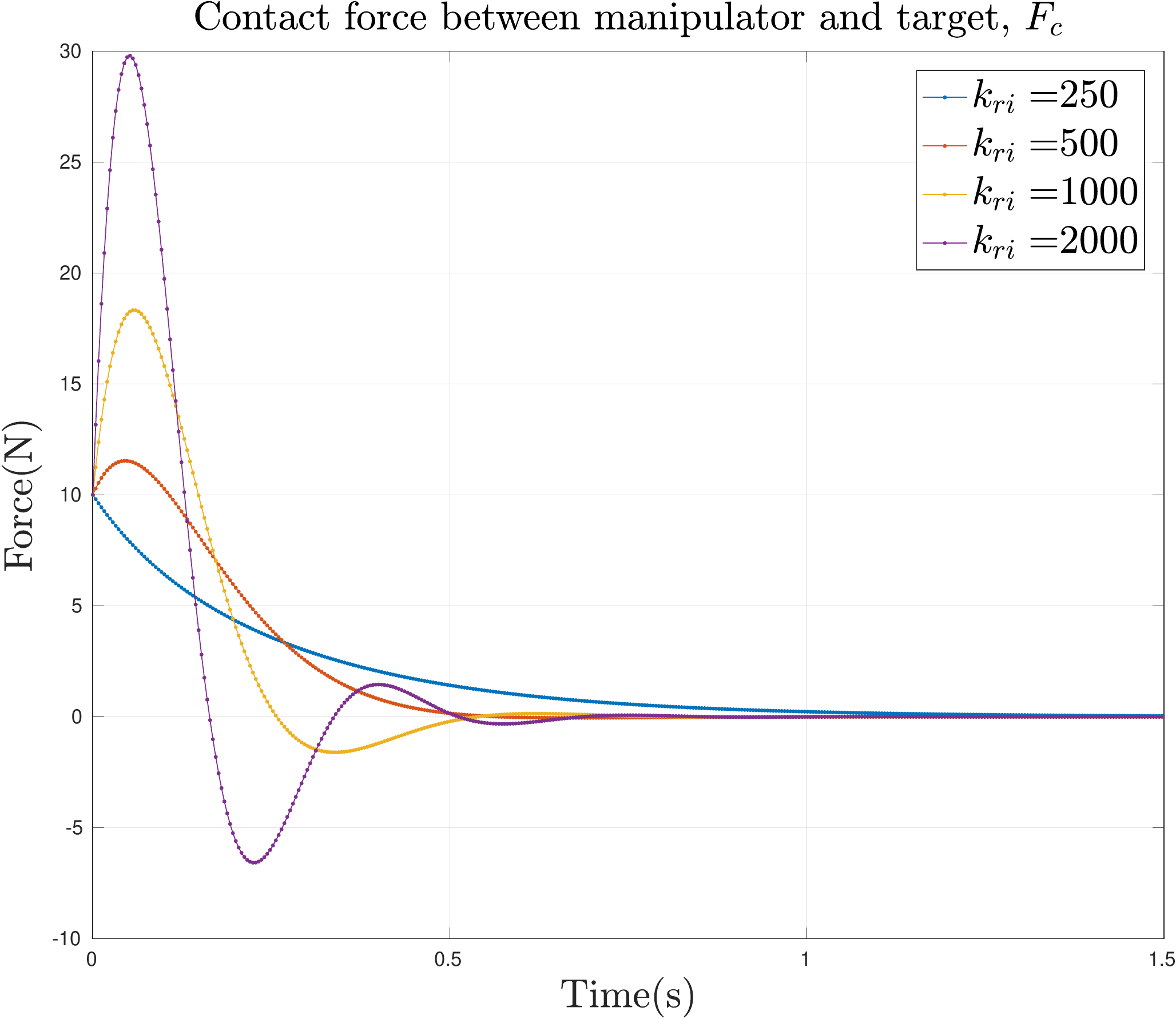} \caption{} \label{fig:sim_res_b} \end{subfigure}\quad \begin{subfigure}{0.3\textwidth} \centering \includegraphics[width=1\textwidth]{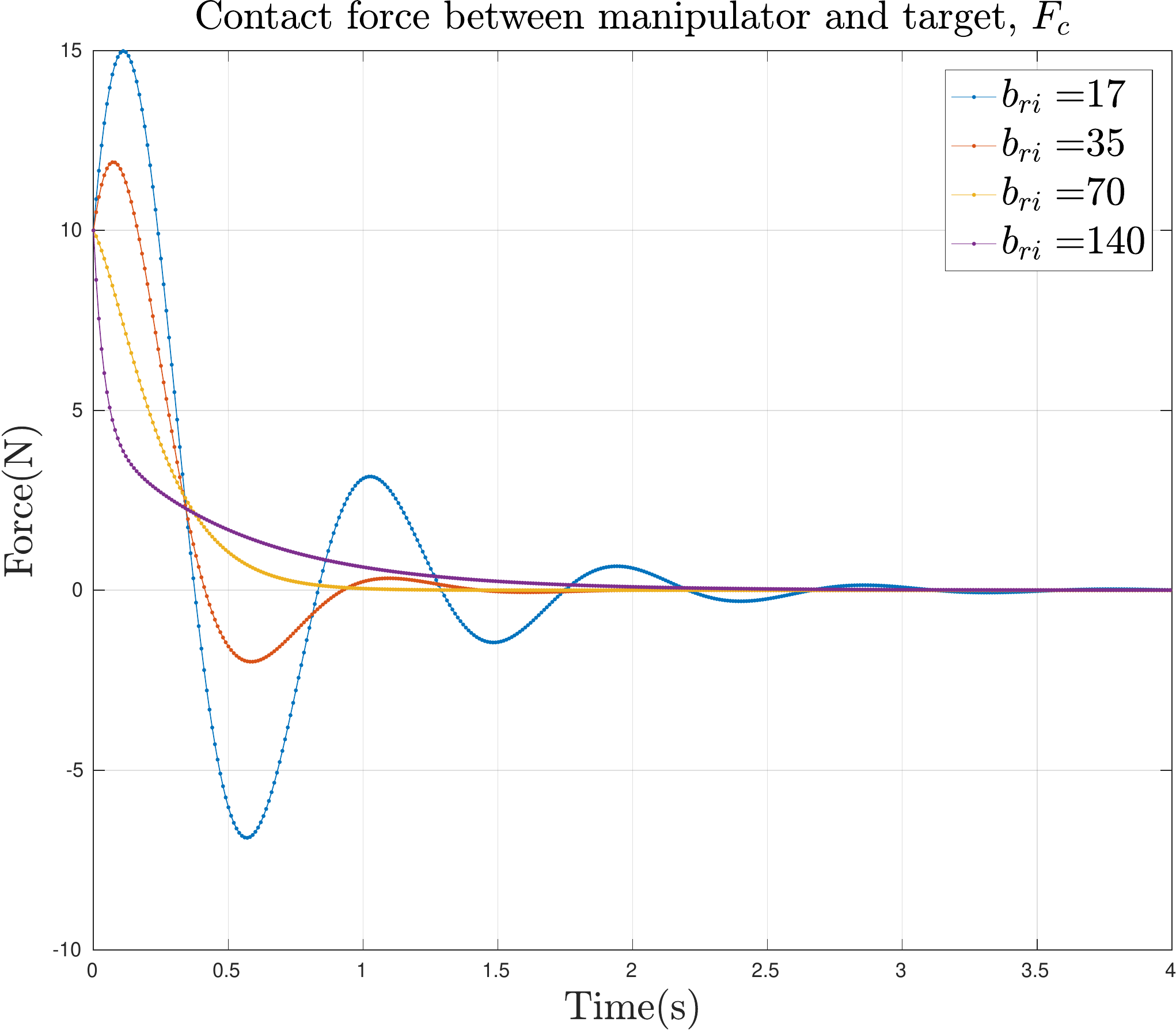} \caption{} \label{fig:sim_res_c} \end{subfigure}
    }
    \caption{Simulation results for contact force($F_c$) between manipulator and free-floating object: (a) varying mass ratio ($m_{ri}/m_t$), (b) varying stiffness($k_{ri}$), and (c) varying damping parameter($b_{ri}$) of manipulator arm}
    \label{fig:sim_res}

 \end{figure*}

Fig. \ref{fig:sim_res} shows the simulation results for contact force $F_c$; the three sub-figures are generated by solving (\ref{control_sim_man}) with varying mass ratio, stiffness and damping parameters. Parameters for simulation are listed in Table \ref{sim_table}. At the start of simulation, the target is at rest and the robot has an approach velocity. It is assumed that the simulation starts in contact, i.e. impact phase is not considered here. As the simulation proceeds, the robot and target achieve same velocity while maintaining contact. The objective of the simulation with varying mass ratio between manipulator and object was to study the effect of relative inertia on interaction force (Fig. \ref{fig:sim_res_a}). The target is difficult to move if the inertia of the target with respect to the manipulator is high. Also in this case the final velocity of the target and manipulator is lower which reduces the chase length and duration. As seen from Fig. \ref{fig:sim_res_a}, high relative inertia results in higher impact and hence contact break. Consequently, lower relative inertia results in damped oscillation in transient behavior, and manipulator continues to be in contact with target. 

\begin{table}[h!]
\begin{center} 
 \textbf{Initial conditions}
 \vspace{2mm}

 \begin{tabular}{||c c c||}
 \hline
 Manipulator velocity[m/s] & $\dot{x}_i$ & 0.5 \\
 \hline
 Target velocity[m/s] & $\dot{x}_t$ & 0 \\
 \hline
 Penetration depth[m] & $y(0)$ & 0 \\
 \hline
 Rate of penetration[m/s] & $\dot{y}(0)=\dot{x}_t-\dot{x}_i$ & 0 \\
 \hline
 \end{tabular}

 \vspace{2mm}
 \textbf{Simulation parameters}
 \vspace{2mm}

 \begin{tabular}{||c c c c c||}
 \hline
 Name & Variable & $sim1$ & $sim2$ & $sim3$ \\ [0.5ex]
 \hline
 Target Mass[Kg] & $m_t$ & var & 15 & 25  \\ 
 \hline
 Contact Stiffness[N/m] & $k_c$ & 250 & 250 & 250 \\ 
 \hline
 Contact Damping[Ns/m] & $b_c$ & 20 & 20 & 20 \\ 
 \hline
 Manipulator Mass[Kg] & $m_{ri}$ & var & 7.5 & 7.5   \\
  \hline
 Manipulator Stiffness[N/m] & $k_{ri}$ & 0 & var & 0  \\
  \hline
 Manipulator Damping[Ns/m] & $b_{ri}$ & 100 & 100 & var \\
  \hline
 Friction force[N] & $F_{friction}$ & 0 & 0 & 0  \\
  \hline
\end{tabular}
\end{center}
\caption {Initial conditions and parameters used for simulation. The labels $sim1$, $sim2$, and $sim3$ correspond to Fig. \ref{fig:sim_res_a}, \ref{fig:sim_res_b}, and \ref{fig:sim_res_c}, respectively.} \label{sim_table} 
\end{table}

The objective of simulation with varying stiffness and damping of the manipulator was to study the effect of compliance on interaction force (Fig. \ref{fig:sim_res_b} and \ref{fig:sim_res_c}). The higher the stiffness, the higher the contact force between manipulator and object during transient phase (Fig. \ref{fig:sim_res_b}). High stiffness also results in breaking contact, as seen from negative value of contact force. Similarly, higher value of damping leads to faster decay in oscillations for contact forces, as shown in Fig. \ref{fig:sim_res_c}. 

It can be concluded from these observations that low manipulator to object mass ratio, low stiffness, and high damping are desirable parameters to have for manipulation on free-floating object. As described before, we have assumed that the stiffness and damping are the structural properties of the manipulator and they form passive compliance. Although, stiffness and damping can be actively controlled by impedance controller. Nevertheless, we are not using impedance controller due to the contact breaking problem discussed in experiment section. In the following section, the proposed method focuses on reducing the apparent inertia of manipulator with respect to the free-floating object. We focus on solving the most difficult case shown in Fig. \ref{fig:sim_res_a}, where the mass ratio between manipulator and target $(m_{ri}/m_t)$ is high and thus the target may float away if too much force is applied.

\section{METHOD}
\label{METHOD}

The proposed approach uses force control for maintaining constant minimum force along the motion direction and apparent reduction of manipulator inertia along remaining DOF. A number of DOFs are utilized for maintaining contact and the rest are compliant with measurable contact force. A proportional-integral (PI) controller is used as the constant force controller. For apparent reduction in translational inertia along an axis, measured contact force along this axis is scaled up and applied as additional actuator force. Similarly, for apparent reduction in rotational inertia around an axis, the measured contact torque around this axis is scaled up and applied as additional actuator torque. Using the scaled contact wrench as the actuator wrench emulates the behavior of the target being much heavier than the robotic manipulator. As force control is not feasible before contact phase, impedance control is used for the motion in free space. 

\subsection{Control Law}

A manipulator with open kinematic chain structure with \textit{n} joints is considered. The dynamics of the end-effector in an operational or end-effector space (set x of m independent configuration parameters) is given by

\begin{equation} \label{Man_dyn_Model_oper_1}
M_{ro}(x)\ddot{x}+C_o(x,\dot{x})+G_o(x)+F_c=F
\end{equation}

\noindent
where $M_{ro}(x)$ is $m\times m$ pseudo mass/inertia matrix, $C_o(x,\dot{x})$ is $m\times 1$ Coriolis/centrifugal term, and $G_o(x)$  is $m\times 1$ gravity term   \cite{Khatib1987}. $M_{ro}(x)$ becomes the true inertia matrix when m = n and the manipulator is at a non-singular configuration. $F$ is the generalized wrench at the end-effector and $F_c$ is the contact force. The control force F in (\ref{Man_dyn_Model_oper_1}) can be decomposed to provide a decoupled control structure 

\begin{equation} \label{Man_dyn_Model_oper_1_decouple1}
F=\widehat{M}_{ro}(x)M^{-1}_dF^*+\widehat{C}_o(x,\dot{x})+\widehat{G}_o(x)+\widehat{F}_c
\end{equation}
\noindent
where the circumflex ($\ \widehat{}\ $) denotes estimates of the quantities from (\ref{Man_dyn_Model_oper_1}); $F^*$ is the control or command vector for the decoupled control, and $M_d$ is the desired inertia matrix. With perfect nonlinear compensation and dynamic decoupling (i.e. $\widehat{M}_{ro}(x)= M_{ro}(x)$, $\widehat{C}_o(x,\dot{x}) = C_o(x,\dot{x})$, and $\widehat{G}_o(x) = G_o(x) $) the end-effector is equivalent to a body with inertia $M_d$ in an m-dimensional operational space. 

The control vector $F^*$ in (\ref{Man_dyn_Model_oper_1_decouple1}) can be decomposed along orthogonal DOFs for maintaining contact and apparent inertia reduction. The motion direction gives the direction of force control (chosen by matrix $W$), and remaining orthogonal DOF are the directions for reduced apparent inertia (chosen by matrix $I-W$, $I$ being identity matrix). The control vector is given by  

\begin{equation}\label{f_decouple1}
\begin{aligned} 
F^*=W\left(K_{pFB}(\widehat{F}_c-F^{ref})+K_{iFB}\int_{0}^{t}(\widehat{F}_c-F^{ref})dt\right)\\+(I-W)K_{RI}\widehat{F}_c
\end{aligned}
\end{equation}

\noindent
where $K_{pFB}$ and $K_{iFB}$ are the proportional and integral gain matrix for the PI negative feedback controller used for force control in motion direction; $\widehat{F}_c$ is the measured contact wrench constituting of contact force and torque; $F^{ref}$ is desired interaction wrench and $K_{RI}$ is positive feedback gain matrix for apparent inertia reduction. This composition of $F^*$ results in decoupled second order equations in both the force and reduced apparent inertia directions, i.e.

\begin{equation}
\ddot{x}_{ri}=M^{-1}_d(I-W)\{K_{RI}\widehat{F}_c\} \\
\end{equation}

\begin{equation}
\begin{aligned}
\ddot{x}_f=M^{-1}_dW\{K_{pFB}(\widehat{F}_c-F^{ref})+K_{iFB}\\ \qquad\int_{0}^{t}(\widehat{F}_c-F^{ref})dt\}
\end{aligned}
\end{equation}

\noindent
where $\ddot{x}_{ri}$ is acceleration in reduced apparent inertia directions and $\ddot{x}_f$ is acceleration in force control direction. In this method a number of robot's DOFs are regarded as force-controlled, whereas the rest are controlled for apparent inertia reduction. This approach provides a dynamically decoupled control system with feedback linearization, and there is more flexibility to choose different control subsystems in the controller design. Instead of matching the inertias in $M_d$, we reduce the inertia of the manipulator to achieve the desired behaviour.

\subsection{Reduced Degrees of Freedom Case}
\begin{figure*}[ht]
    \centering

    \makebox[\linewidth][c]{
    \begin{subfigure}{0.2\textwidth} \centering \includegraphics[width=1\textwidth]{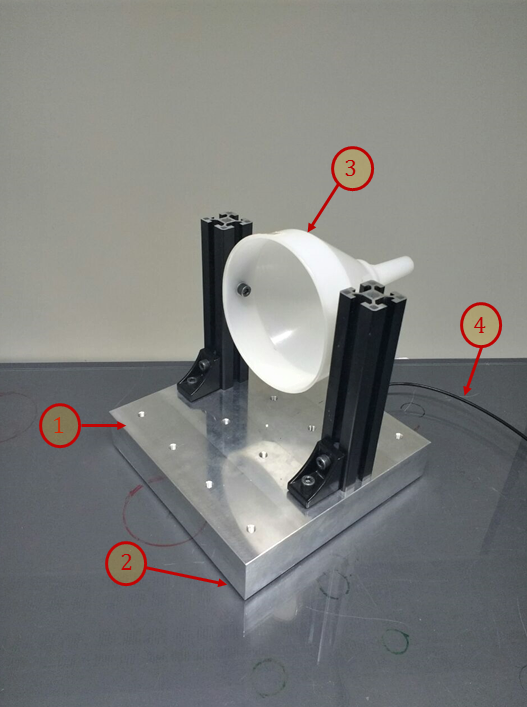} \caption{} \label{fig:setup_target} \end{subfigure} \qquad \begin{subfigure}{0.35\textwidth} \centering \includegraphics[width=1\textwidth]{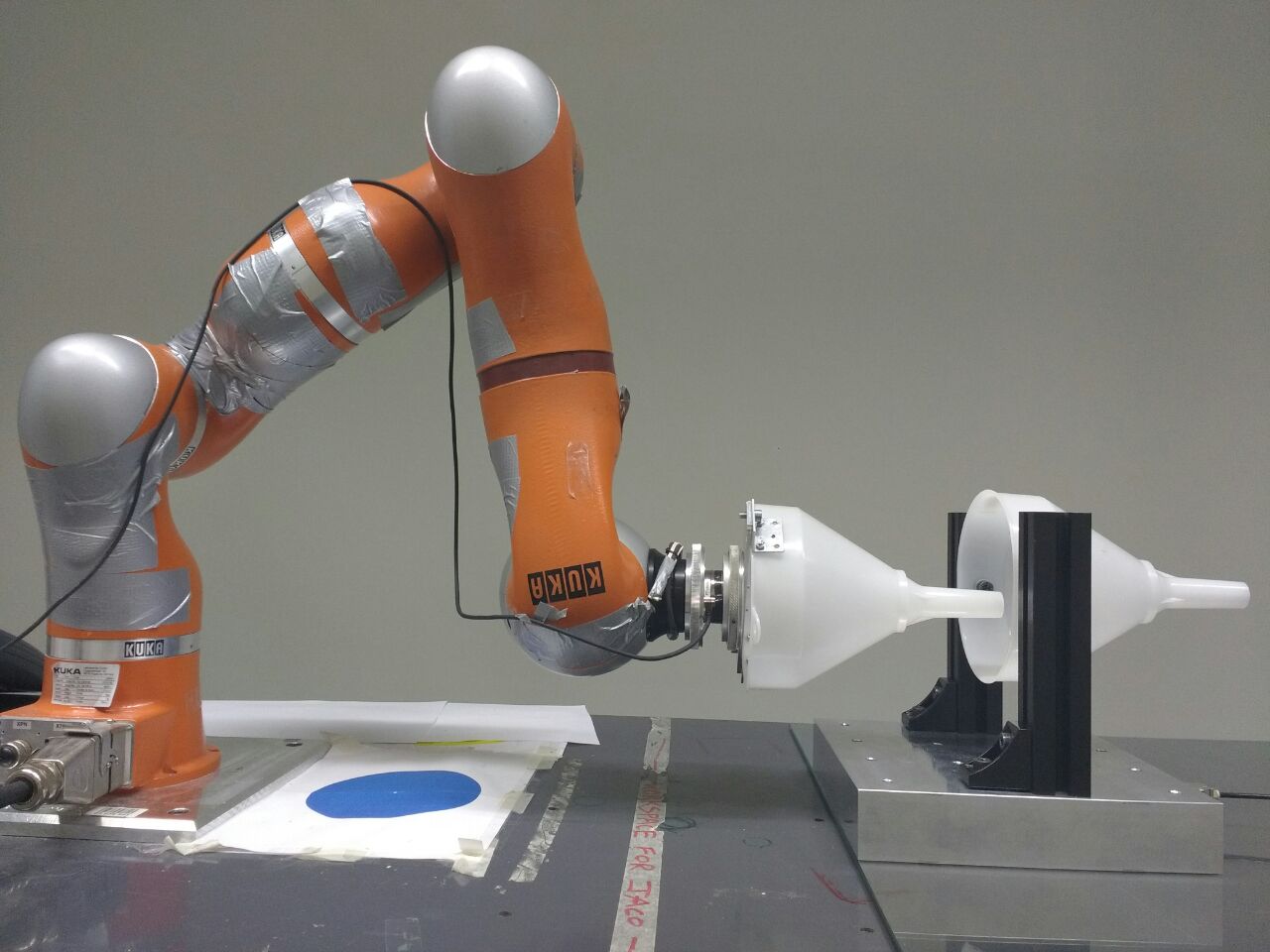} \caption{} \label{fig:setup_r_t} \end{subfigure} \qquad \begin{subfigure}{0.35\textwidth} \centering \includegraphics[width=1\textwidth]{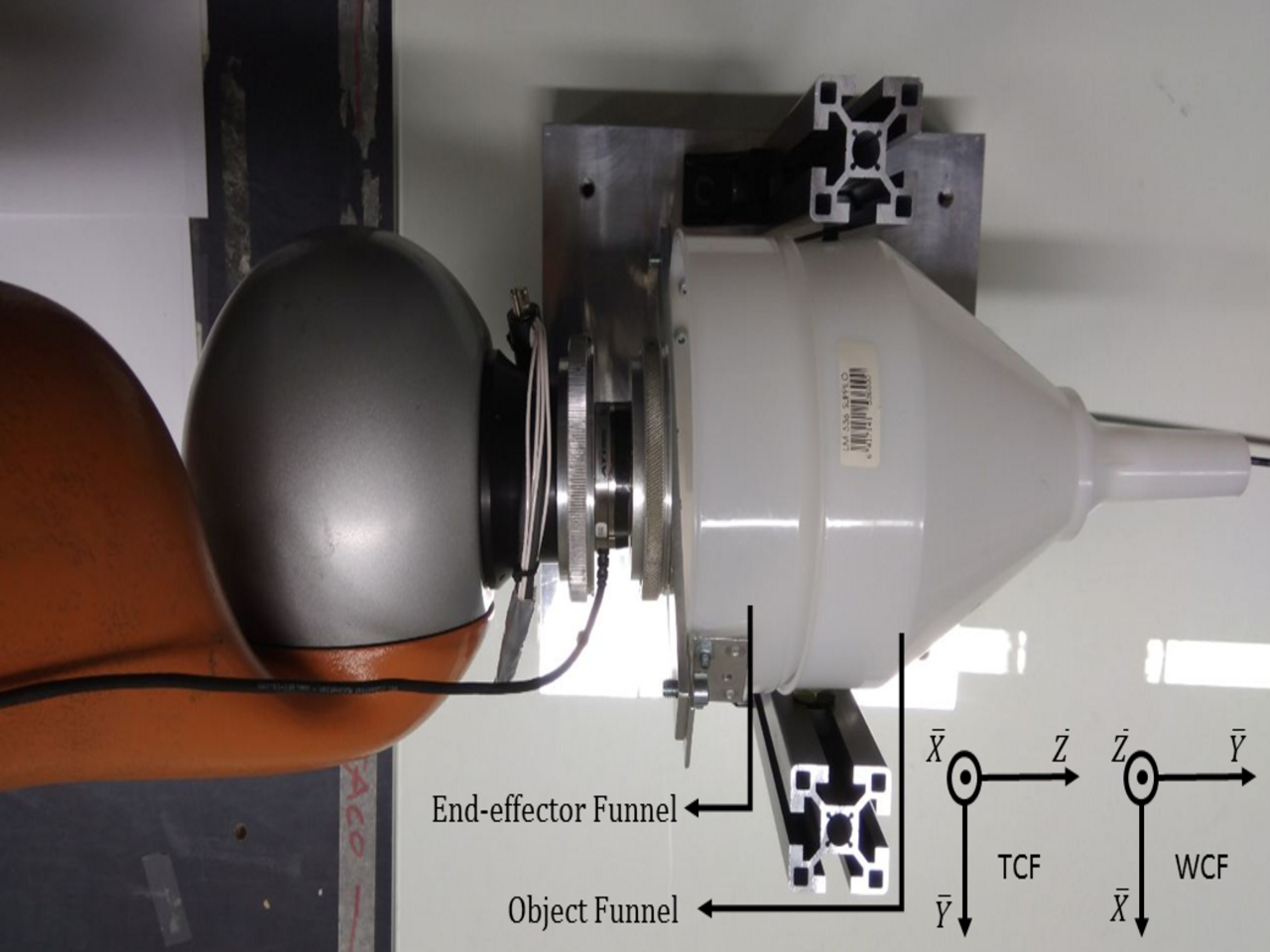} \caption{} \label{fig:setup_coord} \end{subfigure}
    }
    
    \caption{Experimental setup: (a) Free-floating object with (1) Aluminium base with air channel , (2) Air bed, (3) Funnel, (4) Hose for air supply for air bed; (b) General setup for experiments with KUKA LWR4+ robotic manipulator; (c) Experimental setup top view, aligned tools(end-effector funnel and free-floating object funnel) and coordinate systems}

    \label{fig:setup}
 \end{figure*}

The proposed approach can be explained with respect to the experimental setup used in this paper for reduced DOF (Fig. \ref{fig:setup_coord}). The target is assumed to be free-floating on an air bed, hence it has two degrees of translational freedom (Y and Z in Tool Coordinate Frame, TCF) and one degree of rotational freedom (X in TCF). All the axes description used in the following text belongs to TCF. Chosen motion direction or force control direction is along Z-axis, apparent reduction in translational inertia is along Y-axis, and apparent reduction in rotational inertia is around X-axis. The values of matrices from (\ref{f_decouple1}) for reduced-DOF case are

\begin{equation}
W=
\begin{bmatrix}
    1 & 0 & 0 \\
    0 & 0 & 0 \\
    0 & 0 & 0 
\end{bmatrix}, \widehat{F}_c=
\begin{bmatrix}
    f_z \\
    f_y \\
    \tau_x      
\end{bmatrix}, F^{ref}=
\begin{bmatrix}
    f_z^{ref} \\
    0 \\
    0     
\end{bmatrix}\nonumber
\end{equation}

\begin{equation}\label{prop_controller_matrix}
K_{pFB}=
\begin{bmatrix}
    k_{zp} \\
    0  \\
    0  
\end{bmatrix}, K_{iFB}=
\begin{bmatrix}
    k_{zi} \\
    0 \\
    0      
\end{bmatrix}, K_{RI}=
\begin{bmatrix}
    0 \\
    k_{yp} \\
    k_{xp}     
\end{bmatrix}
\end{equation}

\vspace{2mm}

\begin{figure}[ht]
  \begin{center}
    \includegraphics[width=1\linewidth]{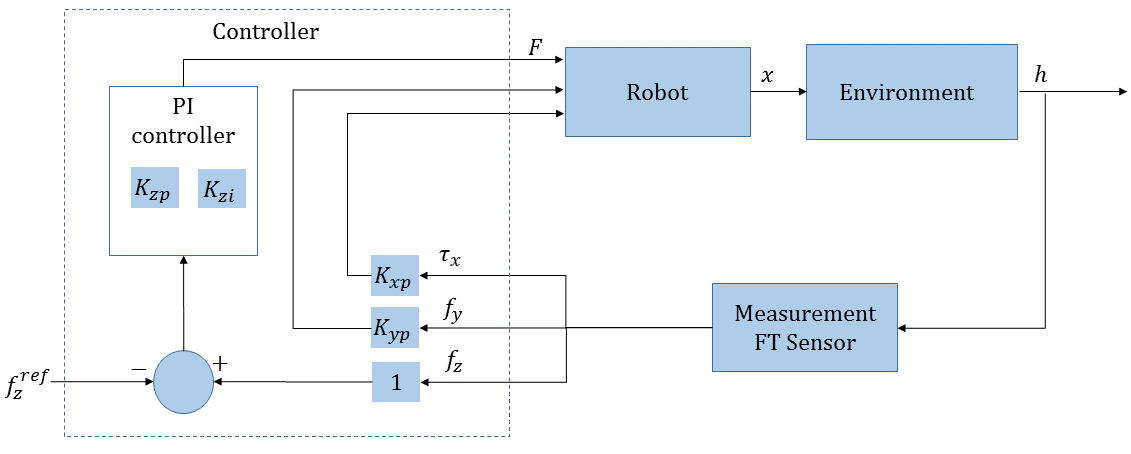}

    \caption{Block Diagram for the proposed controller: Constant force controller with apparent inertia reduction}
    \label{fig:proposed_controller}
  \end{center}
\end{figure}

By substituting matrix values from (\ref{prop_controller_matrix}) in (\ref{f_decouple1}), a 2-DOF form of the proposed controller is deduced as

\begin{equation} \label{prop_controller}
F^*= 
\begin{cases}
    \text{Impedance controller} ,& \text {$F_c =0$}\vspace{2mm}\\
    k_{xp}\tau_x\bar{x}+k_{yp}f_y\bar{y}-\{k_{zp}(f_z-f_z^{ref})\\ \qquad\qquad\qquad+{k_{zi}}\int_{0}^{t}(f_z-f_z^{ref})dt\}\bar{z},              & \text{$F_c \neq0$ }
\end{cases}
\end{equation}

\noindent
where $F_c=0$ is for the free space motion before contact, which is controlled by an impedance controller to reduce the impact. In case of contact, $F_c \neq0$ and we use our proposed controller, the constant force PI controller with apparent inertia reduction. Proportional and integral gain for the controller are $k_{zp}$ and $k_{zi}$, respectively. $k_{xp}$ and $k_{yp}$ are the reaction force gains. $\tau_x$, $f_y$, and $f_z$ are the contact torque around the X-axis, and the contact force along Y and Z axes, respectively. $f_z^{ref}$ is the desired contact force along Z-axis. $\bar{x}$, $\bar{y}$, and $\bar{z}$ are unit vectors along X, Y, and Z-axis. The block diagram for the proposed controller is given in Fig. \ref{fig:proposed_controller}.

 \begin{table}[h!]
\begin{center} 
 
 \textbf{Parameters for external impedance controller}
 \vspace{2mm}

 \begin{tabular}{||c c c c||}
 \hline
 Stiffness(in TCF) & X & Y  & Z \\ [0.5ex] 
 \hline
 External Linear Stiffness[N/m] & 0 & 0 & 500  \\ 
 \hline
 External Angular Stiffness[Nm/rad] & 0 & 0  & 0 \\ 
 \hline\hline
 Damping(in TCF) & X & Y  & Z \\ [0.5ex] 
 \hline
 External Linear Damping in[Ns/m] & 1 & 1 & 1  \\ 
 \hline
 External Angular Damping in[Nms/rad] & 1 & 1 & 1  \\ 
 \hline
\end{tabular}

\vspace{2mm}
\textbf{Constant force controller with apparent inertia reduction}
\vspace{2mm}

 \begin{tabular}{||c c||} 
 \hline
 Proportional gain ($k_{zp}$) & 1 \\ 
 \hline
 Integral gain ($k_{zi}$) & 0.07 \\ 
 \hline
 \end{tabular}

\end{center}
\caption {Controller parameters for experiments.} \label{exp_table} 
\end{table}

\section{EXPERIMENTS AND RESULTS}
\label{EXPERIMENTS}
The experimental scenario of this paper is alignment of workpieces (two similar funnels) using compliant motions as shown in Figure \ref{fig:setup_coord}. 
\subsection{Hardware Setup}
The hardware setup consists of an arm manipulator and a free-floating target.
\subsubsection{Manipulator}
The robot used was 7-DOF KUKA LWR 4+ manipulator with an ATI mini 45 force/torque sensor attached between the flange and the end-effector. One of the funnel shaped guides was attached to the end-effector and the other funnel was mounted on a free-floating object platform (Fig. \ref{fig:setup_target}).

To implement the controller on the robot, we used KUKA's Fast Research Interface (FRI) \cite{Schreiber2010} with control frequency of 200Hz. The control law for the Cartesian impedance control of the KUKA LWR4+ through FRI is 
\begin{equation} \label{internal_impedance_control}
\tau_{cmd}=J^{T}(k_c(x_{cmd}-x_{msr})+D(d_c)+F_{cmd})+f_{dyn}(q,\dot{q},\ddot{q})
\end{equation}
where $J$ is the Jacobian. The control law represents a virtual spring $k_c(x_{cmd}-x_{msr})$. The stiffness of the virtual spring, $k_c$, the damping factor, $d_c$, the desired Cartesian position, $x_{cmd}$, and the
superposed force/torque term, $F_{cmd}$, can be dynamically set. The term $f_{dyn}(q,\dot{q},\ddot{q})$ is the dynamic model of the robot and compensates for gravity torques, coriolis, and centrifugal forces. We implemented the controller through $F_{cmd}$ by setting $k_c = 0$ and $d_c=0$. To negate the effects of gravity, linear motion along X-axis and rotational motion around Y and Z axes were constrained by setting the stiffness in internal impedance controller corresponding to these axes to maximal value. For the other axes, the proposed controller was implemented via $F_{cmd}$.

\subsubsection{Free-Floating Target}
A two-dimensional micro-gravity emulator that floats a target object on an air bed was developed in-house (Fig. \ref{fig:setup_target}). This system can be used to emulate the planar (3-DOF) version of a 6-DOF space robot. The free-floating platform was made of an aluminium block ($25\times25\times5$ cm), with an in-built air channel. This air channel creates an air bed under the target of a few micron thickness. The air pressure is adjustable with an air pressure regulator and an air valve in the range of 2-8 bars. In the experiments, 2 bar pressure was used for making the whole setup float on a glass surface. The floating target weighed 12kg. 

Two primary limiting factors in this kind of setup are the residual friction and rigidity of the air hose. Experiments were performed to estimate the coefficient of friction by using a video camera with a spatial resolution of $1280\times720$ and temporal resolution of 50 frames per second. An impulse force was applied to the free-floating target, putting it in motion. Deceleration of the free-floating platform was calculated by using initial and final velocity estimates from the video. The coefficient of friction of the free-floating target was found to be less than the resolution of the camera setup, which gives an upper bound of 0.01 for the friction coefficient of the free-floating target.

\subsection{Design of Experiments}

Experiments were performed using the experimental setup shown in Fig. \ref{fig:setup_r_t}. The goal of these experiments was to implement and compare three methods of manipulator control: (i) Force control with apparent inertia reduction (proposed approach), (ii) Force control, and (iii) Impedance control. The main focus was to study two properties: first, whether or not the manipulator is able to maintain contact with the free-floating object during manipulation. Second, to evaluate the effect of apparent inertia reduction on minimization of interaction forces. 

At every update interval, the desired position of the end-effector was calculated to make the end-effector follow a straight line along Z-axis, in the YZ plane. While following this trajectory, the funnel on the manipulator interacts with the target funnel, which helps in guiding the manipulator funnel towards alignment. In methods (i) and (ii), impedance controller was only used during the chase phase before contact. As soon as contact was detected, the controller switched to proposed force control with apparent inertia reduction in method (i) and to force control in method (ii), both described in (\ref{prop_controller}). In method (iii), the impedance controller is used throughout the experiment. The parameters for free space impedance control and contact phase force control are given in Table \ref{exp_table}.

An initial set of experiments was performed to find the minimum value of reference or desired contact force ($f_z^{ref}$) along Z-axis with apparent reduction in translational inertia along Y-axis, and apparent reduction in rotational inertia around X-axis. It was found that the end-effector funnel does not slide on the target funnel if the value of $f_z^{ref}$ was below 1N. This can be attributed to the Coulomb friction between funnels and internal friction in the joints of manipulator arm.

\subsection{Compared Methods}

\subsubsection{Method (i), Force Control with Apparent Inertia Reduction}
Two cases were considered for this method: (i) apparent reduction of both translational and rotational inertia (using $k_{xp}$ and $k_{yp}$), and (ii) apparent reduction of only translational inertia (using $k_{yp}$). This was done to understand the effect of reducing apparent inertia along a subset of $(I-W)$ directions in (\ref{f_decouple1}). In both cases interaction forces were minimized by reducing the apparent inertia according to (\ref{prop_controller}). 

Experiments were performed to empirically derive the values of $k_{xp}$ and $k_{yp}$. The theoretical lower limit of $k_{xp}$ and $k_{yp}$ is 0. The theoretical upper limit for $k_{xp}$ is the value at which applied controller torque around X-axis is able to overcome the manipulator inertia around X-axis. Similarly, theoretical upper limit for $k_{yp}$ is the value at which applied controller force along Y-axis is able to overcome the manipulator inertia along Y-axis. 

\subsubsection{Method (ii), Force Control}
Force control is a subset of the proposed controller with $k_{xp}=0$ and $k_{yp}=0$ in (\ref{prop_controller}). As mentioned in the previous subsection, the end-effector funnel did not slide on the target funnel for the value of $f_z^{ref}$ below 1N. By using only force control, higher value of interaction force was required to achieve the alignment. As $f_z^{ref}$ was increased in magnitude, the sliding resulted in alignment of tools. However, this also resulted in the free-floating object moving faster posing a challenge as the manipulator has to chase the target. 

\subsubsection{Method (iii), Impedance Control}
The impedance controller was implemented as 
\begin{equation} \label{external_impedance_control}
F_{cmd} = K_e(x_{current}-x_{setpoint})+B_e\dot{x}
\end{equation}
where $K_e$ and $B_e$ are the stiffness and damping matrix. $x_{current}$ is the current pose of the tool tip, $x_{setpoint}$ is the desired pose, and $\dot{x}$ is the velocity of the end-effector. The actuator force is calculated by impedance controller. To achieve the goal of minimizing interaction forces, low velocity of the manipulator was used with high compliance. The motion is along Z-axis, translation along Y-axis and rotation around X-axis is completely compliant.

\subsection{Results}

\subsubsection{Maintaining contact during manipulation}
\begin{figure*}[ht]
    \centering

    \makebox[\linewidth][c]{
    \begin{subfigure}{0.5\textwidth} \centering \includegraphics[width=0.9\linewidth]{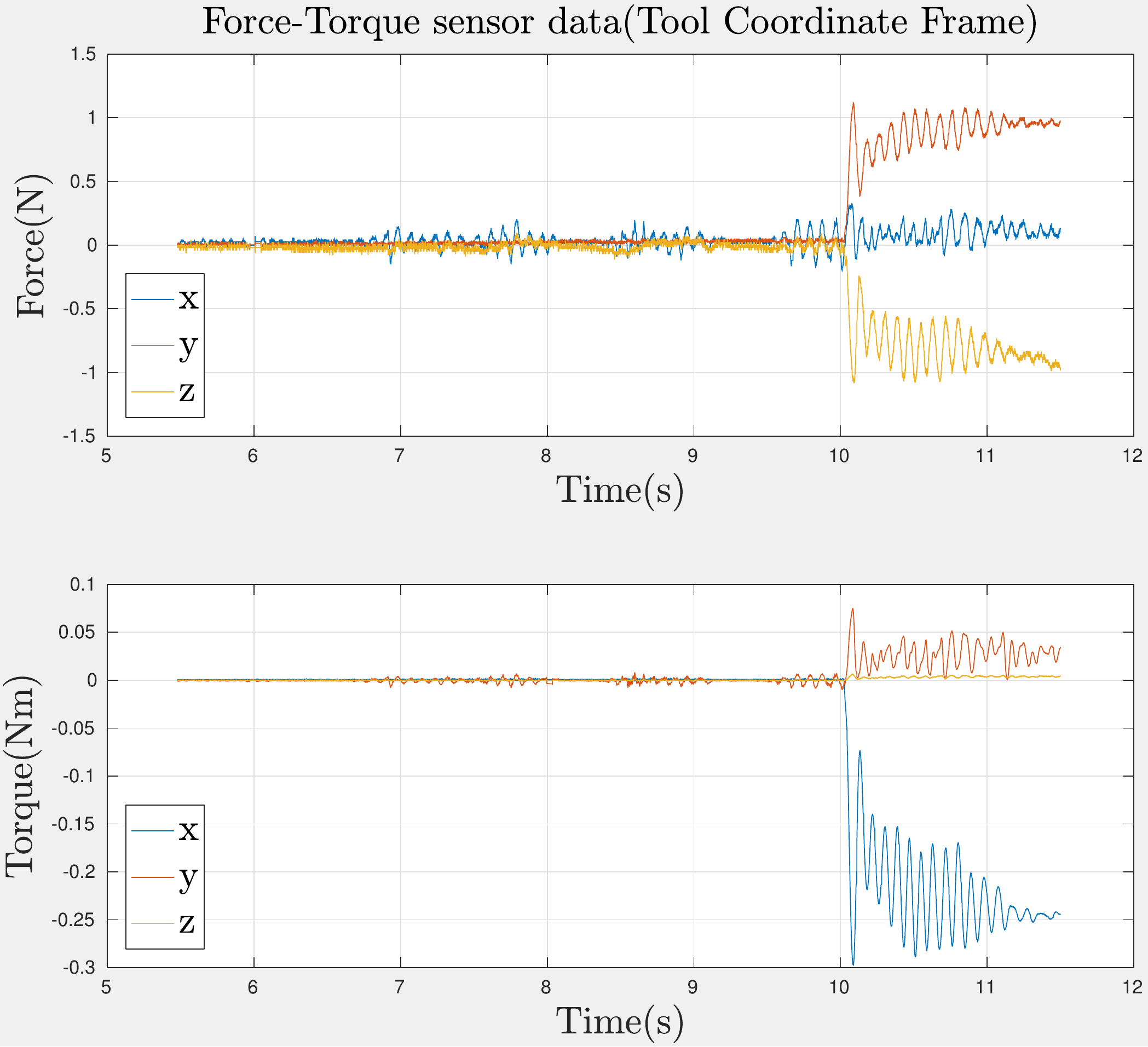} \caption{} \label{fig:exp_fc_tri_a} \end{subfigure} 
    
    \begin{subfigure}{0.5\textwidth} \centering \includegraphics[width=0.9\linewidth]{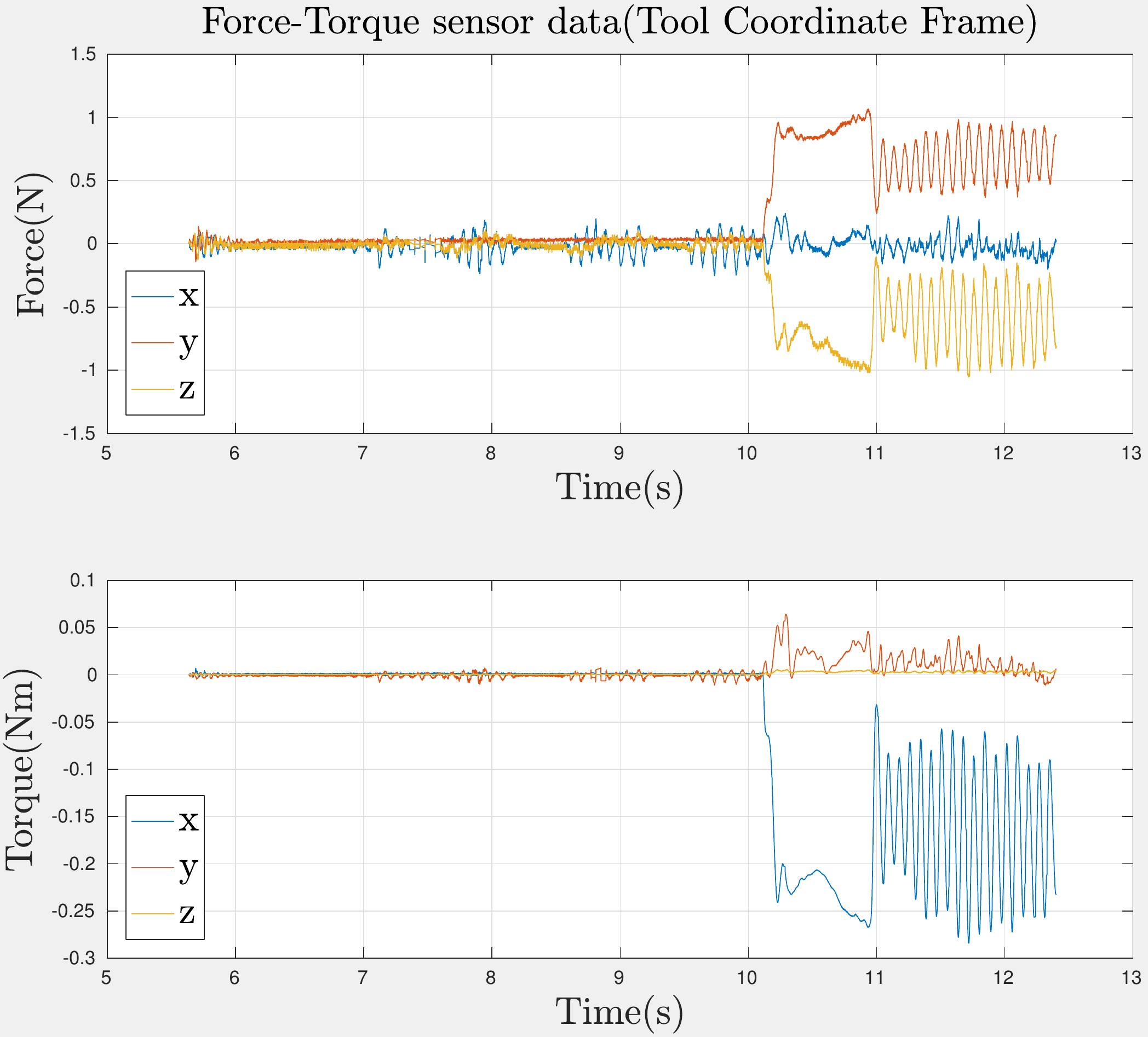} \caption{} \label{exp_fc_ti_1_a} \end{subfigure}
    }
    \makebox[\linewidth][c]{
    \begin{subfigure}{0.5\textwidth} \centering \includegraphics[width=0.9\linewidth]{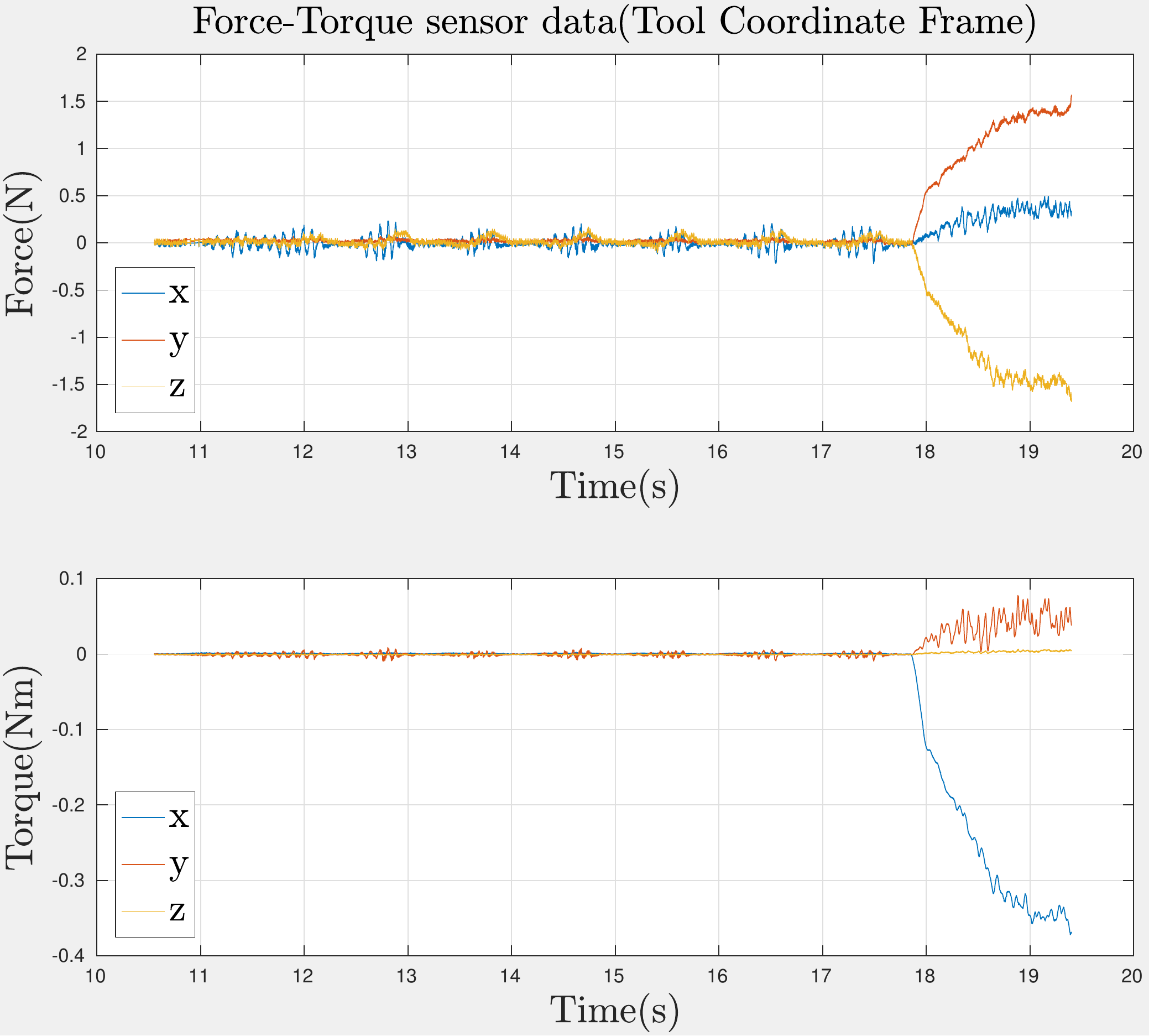} \caption{} \label{exp_fc_a} \end{subfigure}

    \begin{subfigure}{0.5\textwidth} \centering \includegraphics[width=0.9\linewidth]{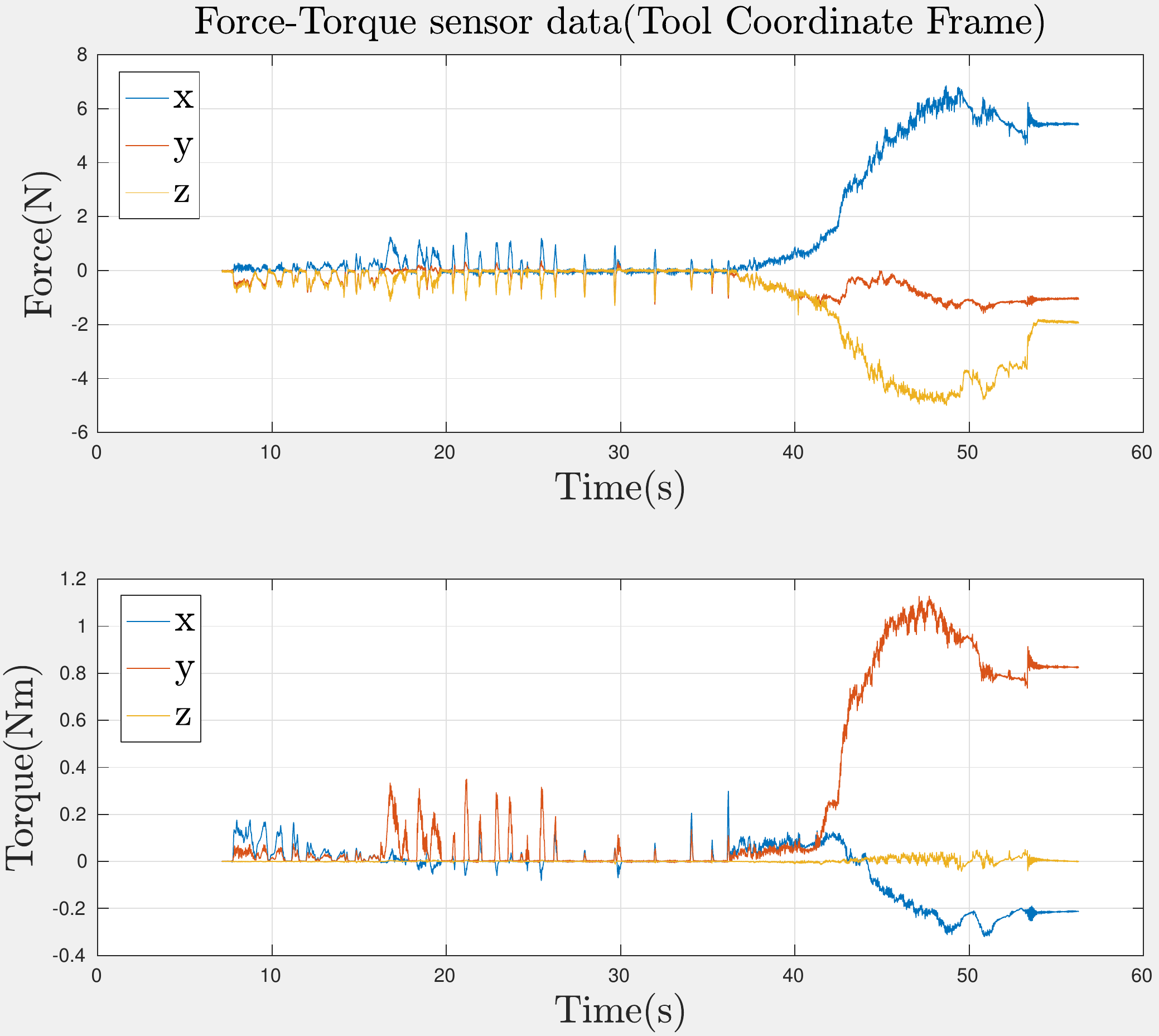} \caption{} \label{fig:exp_imp} \end{subfigure}
    }
    \caption{Force-Torque sensor data showing contact or interaction wrench for the control methods: (a) Method I case I: apparent reduction of both translational and rotational inertia with $f_z^{ref}=0.8$, $k_{xp} = 0.5$, and $k_{yp}=0.2$; (b) Method I case II: apparent reduction of only translational inertia with $f_z^{ref}=0.8$, $k_{xp} = 0$, and $k_{yp}=1$; (c) Method II: Force control with $f_z^{ref}=1.5$; and (d) Method III: Impedance control}

    \label{fig:exp_all_wrench}
 \end{figure*}

Fig. \ref{fig:exp_all_wrench} shows force-torque sensor data for interaction wrench for each control method discussed. Parameters of the methods are shown in Table \ref{res_par}. The experiment started around 8s with the manipulator in impedance control mode. Time of contact varied (observed as rise in contact force), as it depends on the initial placement of free-floating object. Control mode switched at contact for method (i) and (ii), whereas for method (iii) there was no switching of control mode. 

In direct force control approaches, Figs.\ \ref{fig:exp_fc_tri_a}, \ref{exp_fc_ti_1_a}, and \ref{exp_fc_a}, the end-effector funnel slid on the free-floating object funnel resulting in alignment while both were in motion. In the impedance control approach, Fig.\ \ref{fig:exp_imp}, the contact kept breaking. These contact breaks can be seen as small impacts and they can be attributed to the indirect force control where the forces are controlled by the difference between current and desired positions of the manipulator, which tends to fluctuate. This change in forces resulted in a contact bounce problem, that is, the end-effector kept losing contact with the target. Direct force controller worked better for the free-floating target since the contact force could be maintained at a constant minimum force required for the alignment task. Based on this and the simulations in Sec.~III, it can be concluded that the impedance controller is not suitable for alignment if the target is free-floating because of the contact breaking.

\begin{table*}[h!]
\begin{center} 

 \begin{tabular}{||c c c c c c c||} 
 \hline
 Method & $f_z^{ref}$ [N] & $k_{xp}$ & $k_{yp}$ & Experiment start time [s] & Contact time [s] & Experiment end time [s] \\  
 \hline
 Method I case I (Figure \ref{fig:exp_fc_tri_a})& 0.8 & 0.5 & 0.2  & 8 & 10 & 11.5  \\
 \hline
 Method I case II (Figure \ref{exp_fc_ti_1_a})& 0.8 & 0 & 1  & 8 & 11 & 12.5 \\
 \hline
 Method II (Figure \ref{exp_fc_a})& 1.5 & 0 & 0 & 8 & 17.5 & 19.5 \\
 \hline
 Method III (Figure \ref{fig:exp_imp})& - & - & -  & 8 & multiple & 48 \\
 \hline
 \end{tabular}

\end{center}
\caption {Values of parameters corresponding to results shown in Figure \ref{fig:exp_all_wrench}} \label{res_par} 
\end{table*}

\subsubsection{Minimization of Interaction Forces}
Since the impedance controller was unable to maintain contact, the interaction forces were only evaluated for methods (i) and (ii), for which the end-effector funnel slid on the free-floating object funnel resulting in alignment while both were in motion. The reference forces were chosen to smallest values which maintained sliding and contact. 

\begin{figure*}[ht]
    \centering

    \makebox[\linewidth][c]{
    \begin{subfigure}{0.25\textwidth} \centering \includegraphics[width=1\linewidth]{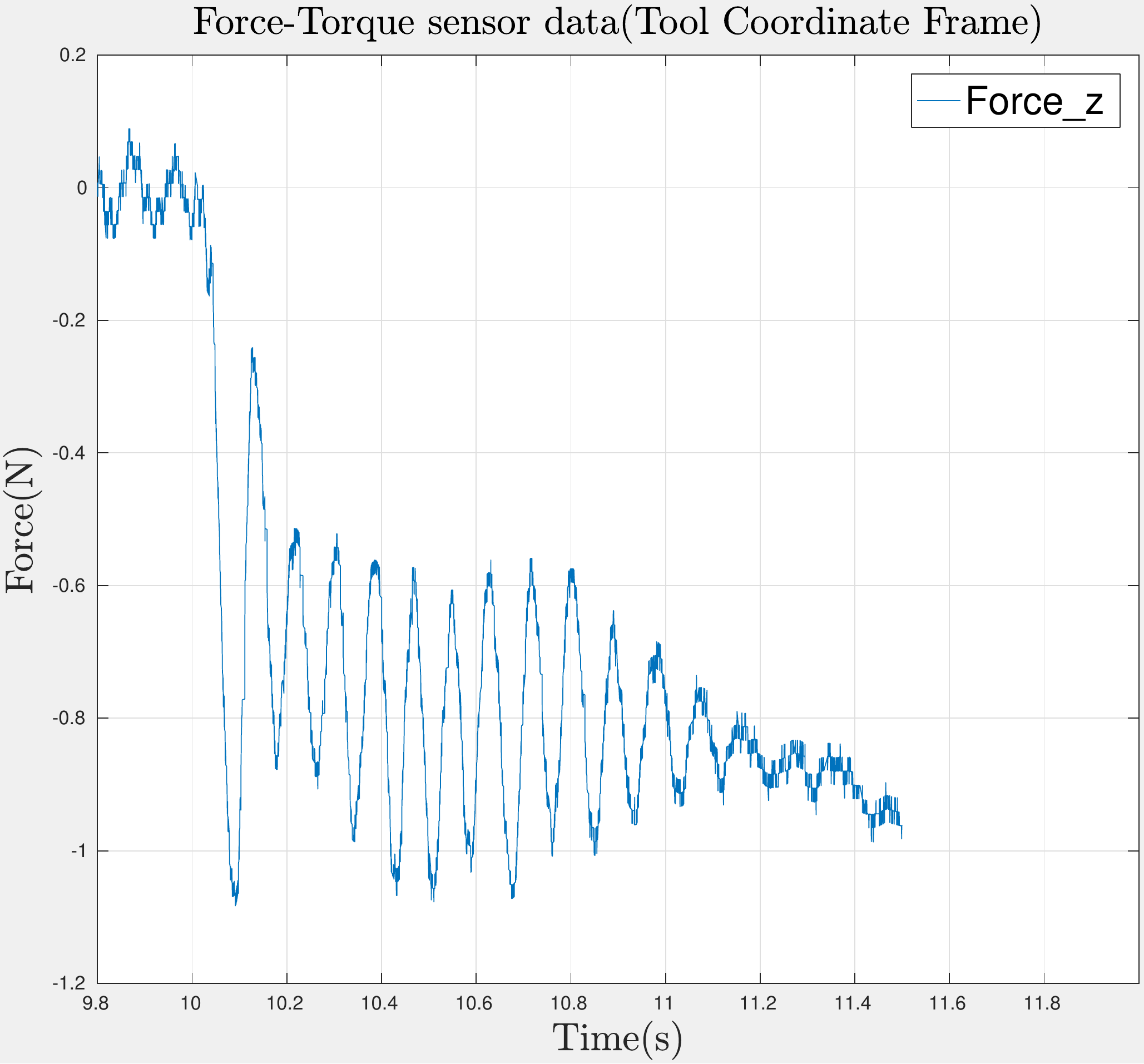} \caption{} \label{fig:exp_fc_tri_b} \end{subfigure} \qquad \qquad \begin{subfigure}{0.25\textwidth} \centering \includegraphics[width=1\linewidth]{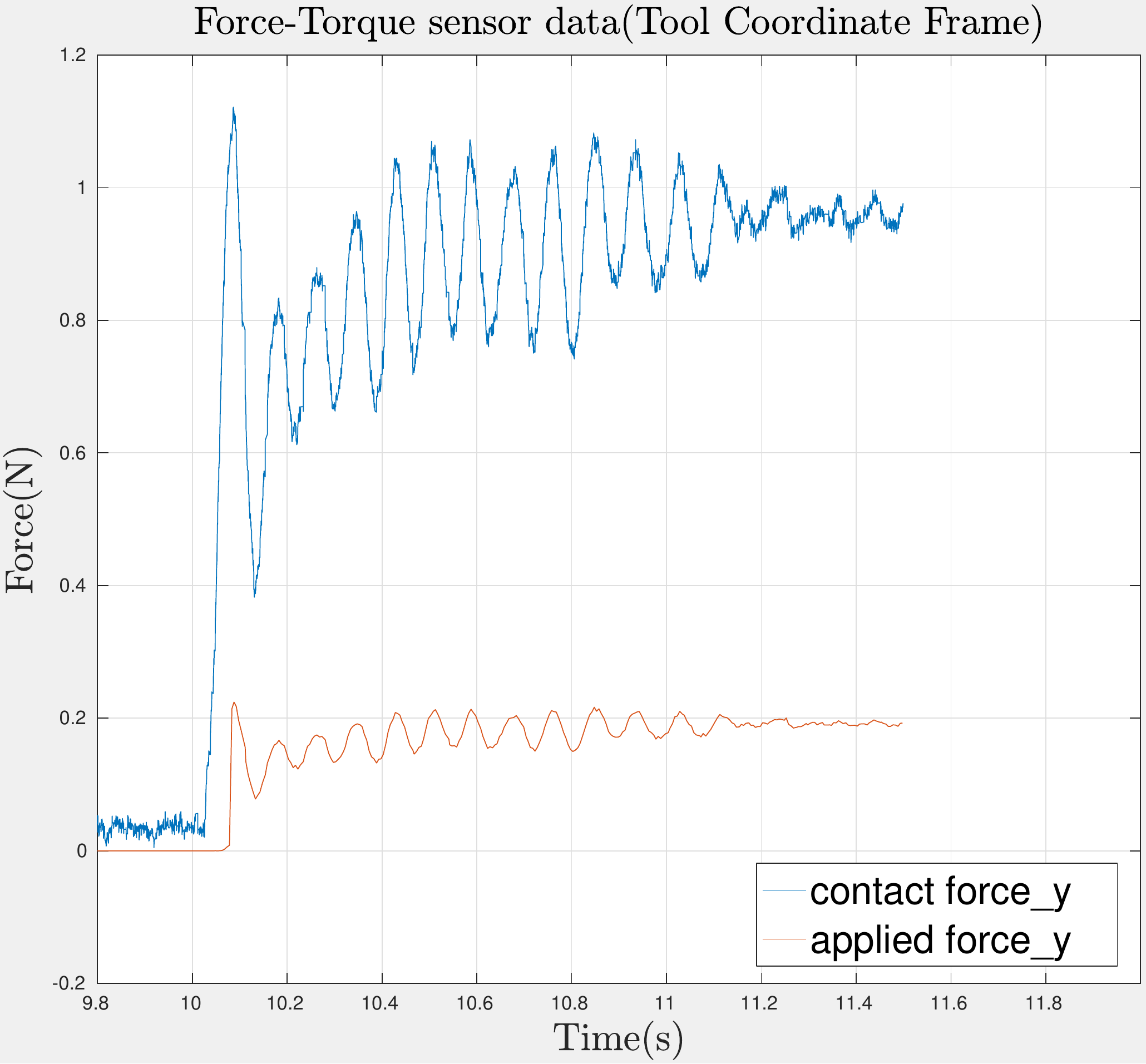} \caption{} \label{fig:exp_fc_tri_c} \end{subfigure}  \qquad \qquad \begin{subfigure}{0.25\textwidth} \centering \includegraphics[width=1\linewidth]{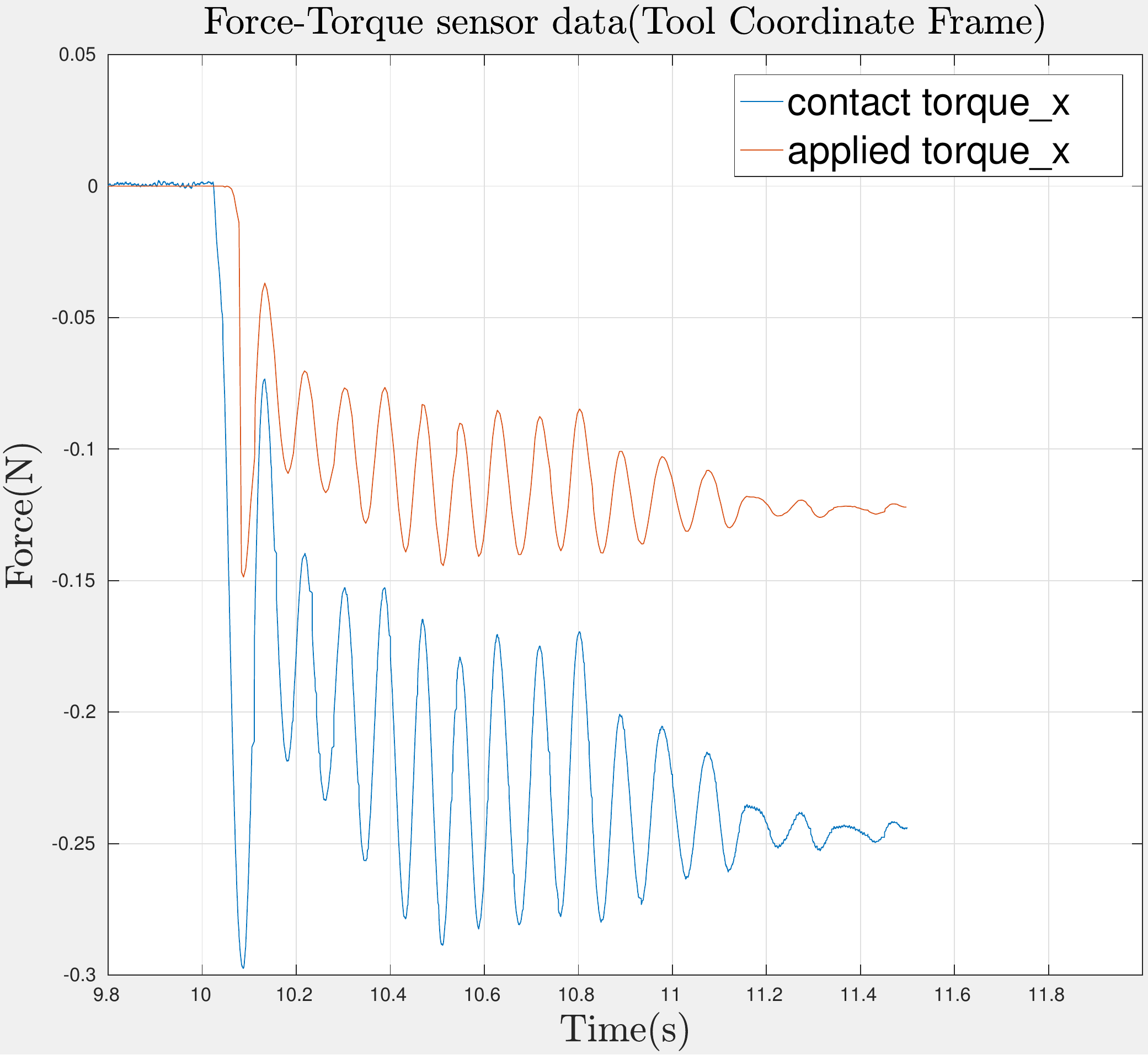} \caption{} \label{fig:exp_fc_tri_d} \end{subfigure} 
    }

    \caption{Results for Method I case I: experiment with force controller and apparent reduction in translational plus rotational inertia of the robotic manipulator for free-floating target, where $f_z^{ref}=0.8$, $k_{xp} = 0.5$, and $k_{yp}=0.2$. (a) Contact force along Z-axis, (b) Contact force and applied force along Y-axis and (c) Contact torque and applied torque around X-axis}

    \label{fig:exp_fc_tri}
 \end{figure*}

\begin{figure}[ht]
    \centering

    \includegraphics[width=0.5\linewidth]{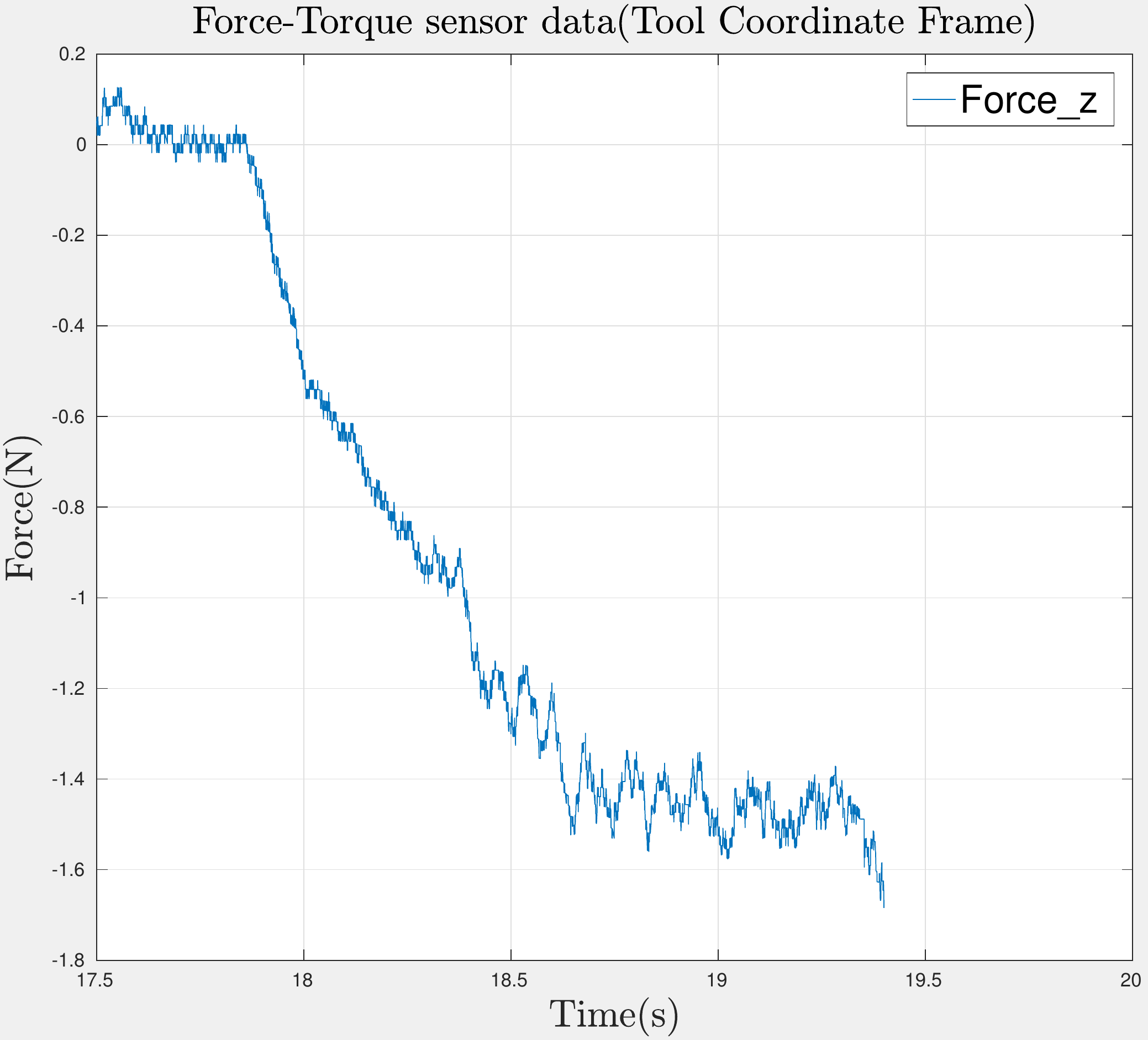}
    
    \caption{Results for Method II: contact force along Z-axis, experiment with force controller, where $f_z^{ref}=1.5$, $k_{xp} = 0$, and $k_{yp}=0$}

    \label{fig:exp_fc}
 \end{figure}

Forces and torques for method (i) are shown in Fig.\ \ref{fig:exp_fc_tri}. It can be observed that the contact force along Z-axis was maintained at $f_z^{ref}=0.8$ (Fig.\ \ref{fig:exp_fc_tri_b}), the applied force for Y-axis was 0.2 times contact force (Fig.\ \ref{fig:exp_fc_tri_c}), and the applied torque for X-axis was 0.5 times contact force (Fig.\ \ref{fig:exp_fc_tri_d}). For method (ii), the reference force needed a higher value of $f_z^{ref}=1.5N$. The measured forces are shown in Fig.~\ref{fig:exp_fc}. It can be seen that the constant force controller is trying to maintain the contact force at $f_z^{ref}=1.5N$. By comparing Fig.~\ref{fig:exp_fc} to Fig.\ \ref{fig:exp_fc_tri_b}, it can be concluded that the proposed inertia reduction allows smaller contact forces to perform the alignment. Hence direct force control with apparent inertia reduction works better for minimizing interaction forces during manipulation.

\begin{figure}[ht]
    \centering

    \makebox[\linewidth][c]{
    \begin{subfigure}{0.25\textwidth} \centering \includegraphics[width=1\linewidth]{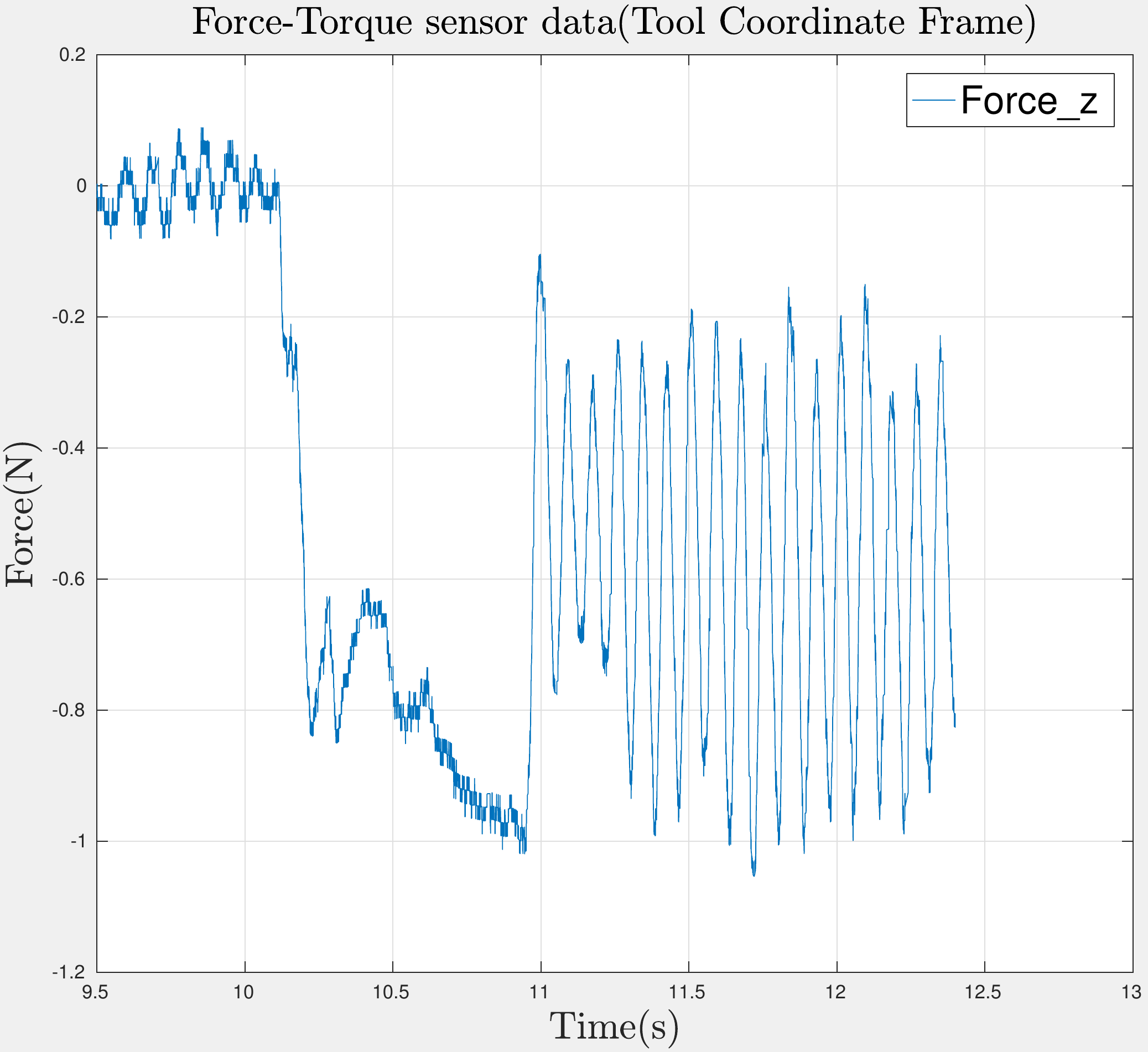} \caption{} \label{exp_fc_ti_1_b} \end{subfigure} \ \begin{subfigure}{0.25\textwidth} \centering \includegraphics[width=1\linewidth]{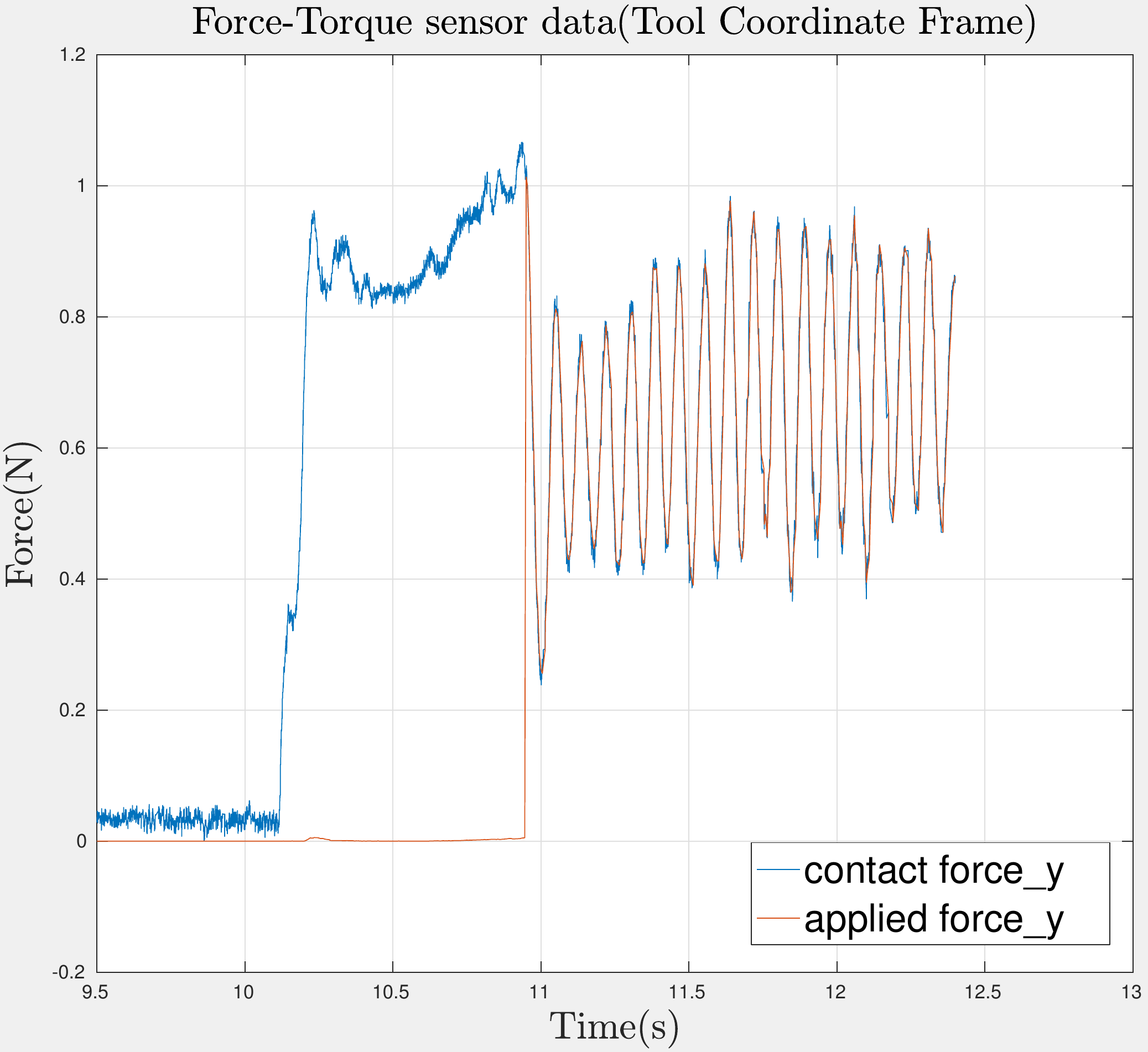} \caption{} \label{exp_fc_ti_1_c} \end{subfigure} 
    }
    \caption{Results for Method I case II: experiment with force controller and apparent reduction in translational inertia of the robotic manipulator for free-floating target, where $f_z^{ref}=0.8$, $k_{xp} = 0$, and $k_{yp}=1$. (a) Contact force along Z-axis, (c) Contact force and applied force along Y-axis}
    \label{fig:exp_fc_ti_1}
 \end{figure}

When both translational and rotational inertia are reduced, it was found experimentally that $k_{xp}$ should be above 0.5 (sliding did not occur at a lower value) and below 2 (contact broke for a higher value). For $k_{yp}$ the lower and upper limits were 0.2 and 1.5, respectively. However, in some applications it can be preferable to reduce the inertia only along a subset of DOF. To study the effect of this option, Fig.\ \ref{fig:exp_fc_ti_1} shows the case where only the translational inertia was reduced. It was found that the minimum contact force required for sliding and alignment remained at $f_z^{ref}=0.8N$. However, the absence of inertia reduction in rotation $k_{xp}=0$, the reduction factor of translational inertia needed to be increased to $k_{yp}=1$ from $k_{yp}=0.2$. It can also be observed that the applied or controller output force for Y-axis is proportional to contact force($k_{yp}=1$) and the ripples from Y-axis force are reflected on the Z-axis force because of the coupling in forces due to the slope of the funnel. The higher value of $k_{yp}$ also results in amplification of ripples which is not desirable for smooth control.

\section{CONCLUSIONS AND FUTURE WORK}
\label{CONCLUSION}

This paper presented a controller for performing compliant manipulation of free-floating objects. The objective of the controller was to minimize the interaction forces while maintaining the contact. The proposed approach achieved this by maintaining constant minimum force along the motion direction and by apparent reduction of manipulator inertia along remaining DOF. The experiments were performed with KUKA LWR4+ manipulator arm and two-dimensional micro-gravity emulator verifying the applicability of the proposed approach. 

Experiments showed that an approach based on direct force control is superior to indirect force control for compliant manipulation of free-floating object. In case of impedance control, the end-effector keeps losing contact with the target. Direct force controller works better for the free-floating target since the contact force can be set to a minimum force required for the alignment task. Furthermore, taking cue from simulations, the apparent inertia of the robotic manipulator was reduced by using the measured contact wrench for additional actuator wrench. This increase in apparent inertia allowed alignment of tools with lower interaction forces.

The proposed control law was verified experimentally for a planar (3-DOF) case. Experimental verification in generic 6-DOF case remains future work. The proposed approach would be also applicable to manipulation of a free floating target by a free flying manipulator, but the interaction of manipulator base control and the reaction forces should be further studied. 

\bibliographystyle{ieeetr}
\bibliography{biblio}

\begin{thebibliography}{10}

\bibitem{HandbookRobotics}
B.~Siciliano and O.~Khatib, {\em Handbook of Robotics}.
\newblock Springer, 2008.

\bibitem{HoganImpedance1985}
N.~Hogan, ``Impedance control: an approach to manipulation: parts i-iii,'' {\em
  Journal of Dynamlc Systems,Measurement, and Control}, vol.~107/1, pp.~1 --
  24, 1985.

\bibitem{Chen2013}
H.~Chen and Y.~Liu, ``Robotic assembly automation using robust compliant
  control,'' {\em Robotics and Computer-Integrated Manufacturing}, vol.~29,
  pp.~293--300, 2013.

\bibitem{Schutter1988}
J.~D. Schutter and H.~V. Brussel, ``Compliant robot motion ii. a control
  approach based on external control loops,'' {\em The International Journal of
  Robotics Research}, vol.~7, pp.~18 -- 33, 1988.

\bibitem{Whitney1977}
D.~Whitney, ``Force feedback control of manipulator fine motions,'' {\em
  Journal of Dynamlc Systems,Measurement, and Control}, vol.~99, pp.~91--97,
  1977.

\bibitem{SalisburyStiffness1980}
J.~Salisbury, ``Active stiffness control of a manipulator in cartesian
  coordinates,'' {\em 19TH IEEE Conference on decision and control},
  pp.~95--100, 1980.

\bibitem{RaibertHybrid1981}
M.~Raibert and J.~Craig, ``Hybrid position/force control of manipulators,''
  {\em Journal of Dynamlc Systems,Measurement, and Control}, vol.~103,
  pp.~126--133, 1981.

\bibitem{Khatib1987}
O.~Khatib, ``A unified approach for motion and force control of robot
  manipulators : the operational space formulation,'' {\em IEEE Journal on
  Robotics and Automation}, vol.~3, pp.~43--53, 1987.

\bibitem{Villani1999}
L.~Villani, C.~Canudas, and B.~Brogliato, ``An exponentially stable adaptive
  control for force and position tracking of robot manipulators,'' {\em IEEE
  Transactions on Automatic Control}, vol.~44, pp.~798--802, 1999.

\bibitem{suomalainen2017}
M.~Suomalainen and V.~Kyrki, ``A geometric approach for learning compliant
  motions from demonstration,'' in {\em Humanoid Robots (Humanoids), 2017
  IEEE-RAS 17th International Conference on}, pp.~783--790, IEEE, 2017.

\bibitem{Yoshida2004}
K.~Yoshida, H.~Nakanishi, H.~Ueno, N.~Inaba, T.~Nishimaki, and M.~Oda,
  ``Dynamics, control and impedance matching for robotic capture of a
  non-cooperative satellite,'' {\em Advanced Robotics}, vol.~18, no.~2,
  pp.~175--198, 2004.

\bibitem{Nakamishi2010}
H.~Nakanishi, N.~Uyama, and K.~Yoshida, ``Virtual mass of impedance system for
  free-flying target capture,'' in {\em 2010 IEEE/RSJ International Conference
  on Intelligent Robots and Systems}, pp.~4101--4106, Oct 2010.

\bibitem{Yoshida2011}
N.~Uyama, D.~Hirano, H.~Nakanishi, K.~Nagaoka, and K.~Yoshida,
  ``Impedance-based contact control of a free-flying space robot with respect
  to coefficient of restitution,'' in {\em 2011 IEEE/SICE International
  Symposium on System Integration (SII)}, pp.~1196--1201, Dec 2011.

\bibitem{Nishida2003}
S.~Nishida and T.~Yoshikawa, ``Space debris capture by a joint compliance
  controlled robot,'' in {\em Proceedings 2003 IEEE/ASME International
  Conference on Advanced Intelligent Mechatronics (AIM 2003)}, vol.~1,
  pp.~496--502 vol.1, July 2003.

\bibitem{Colbaugh1992}
R.~Colbaugh, H.~Seraji, and K.~Glass, ``Impedance control for dexterous space
  manipulators,'' in {\em Proceedings of the 31st IEEE Conference on Decision
  and Control}, pp.~1881--1886 vol.2, 1992.

\bibitem{Schreiber2010}
G.~Schreiber, A.~Stemmer, and R.~Bischoff, ``The fast research interface for
  the kuka lightweight robot,'' in {\em Proc. of the IEEE Workshop on
  Innovative Robot Control Architectures for Demanding (Research) Applications
  – How to Modify and Enhance Commercial Controllers (ICRA 2010)}, 2010.

\end{thebibliography}

\end{document}